\documentclass[journal]{IEEEtran}
\usepackage{amsmath,amsfonts}
\usepackage{algorithmic}
\usepackage{algorithm}
\usepackage{array}
\usepackage[caption=false,font=normalsize,labelfont=sf,textfont=sf]{subfig}
\usepackage{textcomp}
\usepackage{stfloats}
\usepackage{url}
\usepackage{verbatim}
\usepackage{graphicx}
\usepackage{cite}
\usepackage{color}
\usepackage{fancyhdr}

\usepackage[margin=1in]{geometry}
\usepackage{graphicx}
\usepackage{booktabs}
\usepackage{ulem}
\usepackage{multirow}
\usepackage{makecell}
\usepackage[table]{xcolor}
\usepackage{hhline}
\usepackage{algorithm}
\usepackage{algorithmic}
\usepackage{amsmath,amssymb}
\usepackage{caption}
\usepackage{natbib}
\setcitestyle{numbers,square}
\usepackage{footmisc}

\definecolor{tabgray}{gray}{0.9}

\newcommand{\keywords}[1]{\par\textbf{Keywords:} #1\par}

\usepackage{multirow}
\usepackage{color}
\usepackage[driverfallback=dvipdfm,
CJKbookmarks=true, bookmarksnumbered=true, bookmarksopen=true,
colorlinks=true,
citecolor=blue, linkcolor=blue, anchorcolor=red, urlcolor=black
]{hyperref}
\hypersetup{hidelinks,	
            colorlinks=true,	
            allcolors=black,	
            pdfstartview=Fit,	
            breaklinks=true}

\begin{document}

\title{Rethinking Cross-Domain Evaluation for Face Forgery Detection with Semantic Fine-grained Alignment and Mixture-of-Experts}

\author{Yuhan Luo\textsuperscript{*},
Tao Chen\textsuperscript{*},
Decheng Liu     
\noindent
\thanks{Yuhan Luo, Tao Chen and Decheng Liu are with School of Cyber Engineering, Xidian University, Xi’an 710071, Shaanxi, P. R. China (e-mail: dchliu@xidian.edu.cn).}
}



\markboth{Journal of \LaTeX\ Class Files,~Vol.~14, No.~8, Apr.~2026}%
{Shell \MakeLowercase{\textit{et al.}}: A Sample Article Using IEEEtran.cls for IEEE Journals}

\maketitle

\begin{abstract}
Nowadays, visual data forgery detection plays an increasingly important role in social and economic security with the rapid development of generative models. Existing face forgery detectors still can't achieve satisfactory performance because of poor generalization ability across datasets. The key factor that led to this phenomenon is the lack of suitable metrics: the commonly used cross-dataset AUC metric fails to reveal an important issue where detection scores may shift significantly across data domains. To explicitly evaluate cross-domain score comparability, we propose \textbf{Cross-AUC}, an evaluation metric that can compute AUC across dataset pairs by contrasting real samples from one dataset with fake samples from another (and vice versa). It is interesting to find that evaluating representative detectors under the Cross-AUC metric reveals substantial performance drops, exposing an overlooked robustness problem. Besides, we also propose the novel framework \textbf{S}emantic \textbf{F}ine-grained \textbf{A}lignment and \textbf{M}ixture-of-Experts (\textbf{SFAM}), consisting of a patch-level image-text alignment module that enhances CLIP's sensitivity to manipulation artifacts, and the facial region mixture-of-experts module, which routes features from different facial regions to specialized experts for region-aware forgery analysis. Extensive qualitative and quantitative experiments on the public datasets prove that the proposed method achieves superior performance compared with the state-of-the-art methods with various suitable metrics.
The source code is available at \href{https://github.com/Yuhan-Luo/Semantic-Fine-grained-Alignment-and-Mixture-of-Experts}{{\textcolor{blue}{GitHub}}}.
\end{abstract}

\section{Introduction}
\label{sec:intro}

\begin{figure*}[tb]
  \centering
  \includegraphics[width=1.0\linewidth]{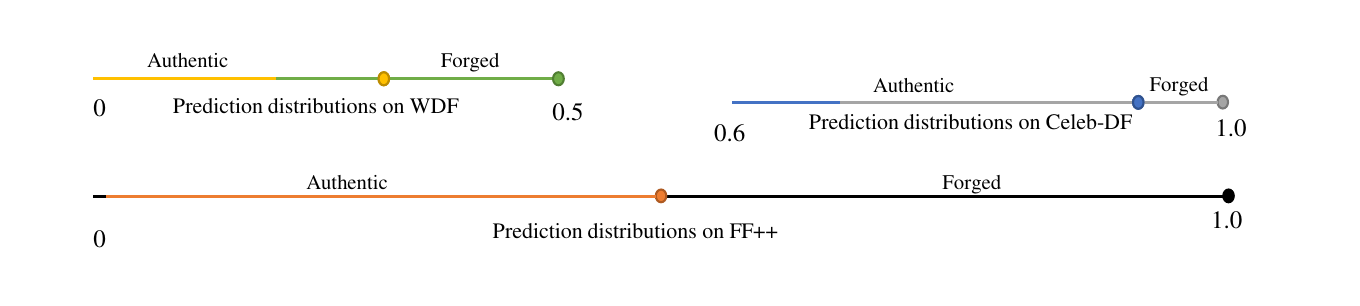}
  \caption{Prediction score distributions of authentic and forged samples across FF++, WDF, and Celeb-DF datasets, showing intra-dataset separability but severe cross-domain distribution shifts.}
  \label{fig:motivation}
\end{figure*}

In recent years, face forgery detection has drawn increasing attention due to its societal and economic implications, and numerous deepfake detection methods\cite{8014963,lyu_exposing_2018,zhao_multi-attentional_2021,9879418} have been developed.
Despite steady progress, reliable deployment in open-world scenarios remains challenging because models often fail to generalize across datasets.
A common practice evaluates cross-domain generalization by training on FF++\cite{Rossler_2019_ICCV} and reporting AUC on other benchmarks. 
However, AUC is invariant to monotonic score transformations and therefore cannot reveal whether detector scores are comparable across domains. 
In practice, dataset-specific biases (e.g., compression pipelines, camera characteristics, post-processing) may cause systematic shifts in the score distribution, leading to inconsistent operating thresholds and unreliable decisions when the test domain is unknown. 
As illustrated in Fig.~\ref{fig:motivation}, \textit{samples can be well separated within each dataset while exhibiting markedly different score ranges across datasets.}

To explicitly measure cross-domain score comparability, we propose \textbf{Cross-AUC}, an evaluation protocol that computes AUC not only within each dataset, but also across datasets by contrasting real samples from one dataset against fake samples from another (and vice versa), and averaging over dataset pairs. 
Benchmarking representative detectors under Cross-AUC reveals substantial drops compared to standard within-dataset AUC, indicating an overlooked vulnerability of current evaluation and training practices.
To improve robustness under Cross-AUC, we further introduce a semantic fine-grained alignment and mixture-of-Experts framework (\textbf{SFAM}), that discourages reliance on dataset-specific shortcuts. 
Specifically, we design a patch-level image-text alignment module (\textbf{PaITA}) that enhances CLIP’s sensitivity to subtle manipulation artifacts, and introduce the facial region mixture-of-experts (\textbf{FaRMoE}), an MoE-style routing mechanism that assigns features from different facial regions to specialized experts for targeted mining of region-specific forgery cues.

The main contributions of our paper are summarized as follows:
\begin{itemize}
    \item We propose a novel specific metric \textbf{Cross-AUC}, which evaluates cross-domain score comparability by computing AUC across dataset pairs. The interesting phenomenon shows that evaluating existing SOTA detectors under the Cross-AUC metric almost reveals substantial performance drops.
    \item We further design the semantic fine-grained alignment and mixture-of-Experts framework (\textbf{SFAM}) to improve the generalization ability with the Cross-AUC metric. The patch-level image-text alignment module can enhance CLIP’s sensitivity to subtle manipulation artifacts, and the facial region mixture-of-experts module aims to assign features from different facial regions to specialized experts for targeted mining of region-specific forgery cues.
    \item Experimental results on the public representative datasets illustrate the superior performance of the proposed SFAM compared with the state-of-the art face forgery detection methods. 
\end{itemize}

\section{Related Work}

\subsection{ViT-based Forgery Detection}
Face forgery detection has evolved from handcrafted feature methods to deep learning-driven paradigms. With the development of convolutional neural networks (CNNs), CNN-based methods became the mainstream of deepfake detection research. Chollet et al.\cite{Chollet_2017_CVPR} proposed Xception, which became the most widely used backbone for early forgery detection; Tan et al.\cite{pmlr-v97-tan19a} further put forward EfficientNet series for better accuracy-efficiency balance, and Lin et al.\cite{10.1007/978-3-031-73016-0_7} applied dynamic data augmentation strategies. However, limited by the inherent local receptive field, CNNs fail to detect high-quality forgeries with subtle artifacts.

\begin{figure*}[tb]
  \centering
  \includegraphics[width=1.0\linewidth]{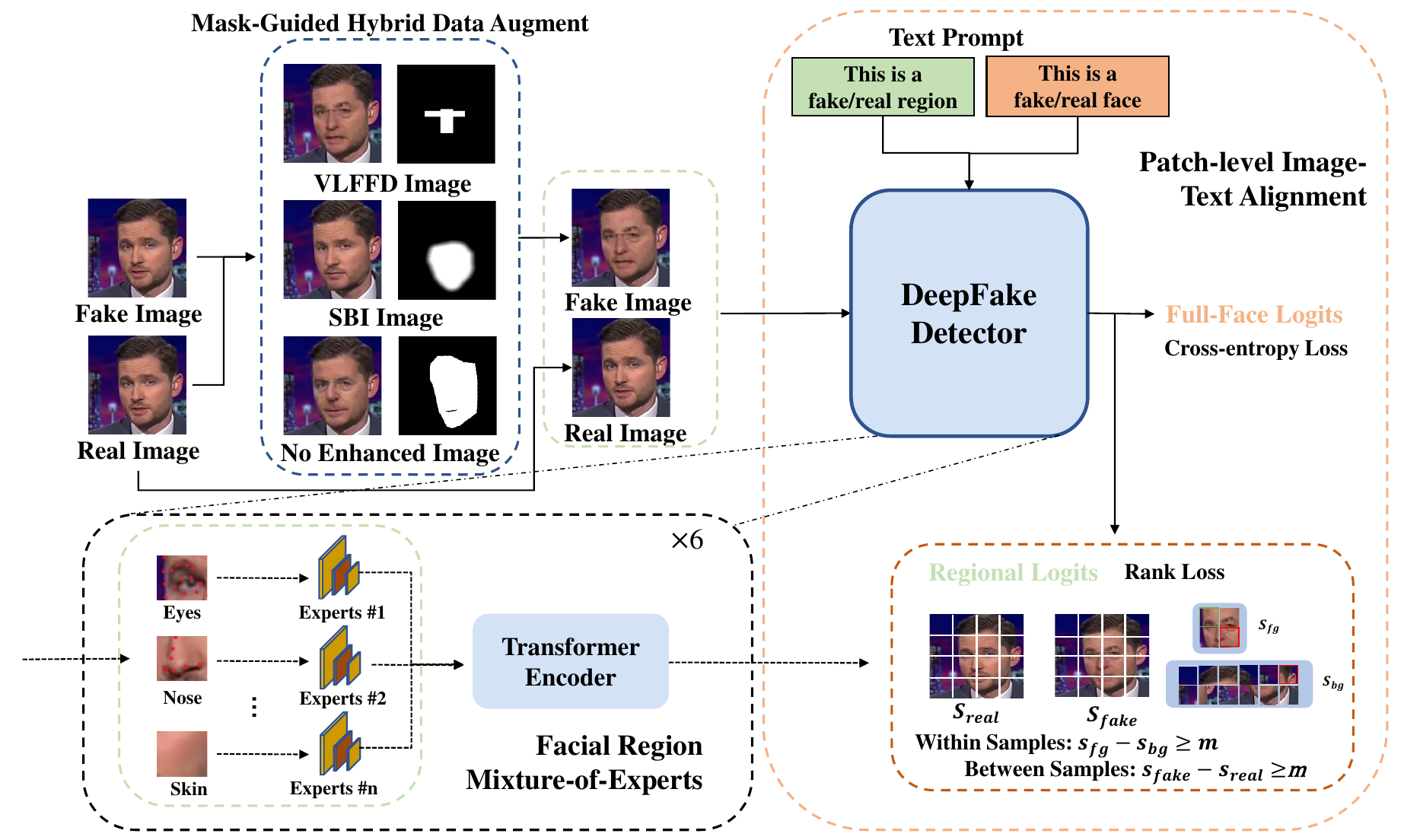}
  \caption{The workflow of the proposed SFAM framework.}
  \label{fig:structure}
\end{figure*}

To address this limitation, Vision Transformer (ViT)\cite{DBLP:journals/corr/abs-2010-11929} has gradually become the core backbone paradigm for face forgery detection. Subsequent studies optimized ViT for forgery detection tasks: Chen et al.\cite{chen2021local} proposed a local relation learning framework to fuse RGB and frequency domain features; Luo et al.\cite{Luo_2021_CVPR} devised a generalizable model to enhance high-frequency forgery feature capture.
In recent years, cross-modal learning represented by CLIP\cite{pmlr-v139-radford21a} has become a key research direction, bringing breakthroughs to the interpretability and generalization of forgery detection. Khan et al. first introduced adaptive prompt tuning into CLIP for universal deepfake detection; Zhang et al.\cite{11023997} designed fine-grained text prompts to help CLIP focus on subtle forgery artifacts; Zhang et al.\cite{10.1007/978-3-031-73223-2_22} constructed the DD-VQA dataset to train a BLIP-based model for both authenticity judgment and interpretable explanation.  Meanwhile, works\cite{sun2023evaclipimprovedtrainingtechniques,10.1007/978-3-031-72983-6_18,NEURIPS2023_6fa4d985} have been done to improve CLIP. These works verified the potential of ViT models, while existing methods still lack targeted optimization for facial regional heterogeneous feature extraction and semantically fine-grained detection.

\subsection{Evaluation Metrics in Deepfake Detection}
Evaluation metrics are the core guidance for the development of face forgery detection technologies. Early research directly adopted general metrics including Accuracy, Precision, Recall and F1-score, which remain auxiliary indicators in most studies. However, due to the widespread class imbalance in forgery detection datasets, these threshold-sensitive metrics cannot comprehensively reflect the model's overall detection ability.

To solve this problem, the Area Under the Receiver Operating Characteristic Curve (AUC) has gradually become the core evaluation metric in this field. AUC comprehensively quantifies the model's ability to distinguish fake and real samples with strong robustness to imbalanced datasets. Rossler\cite{Rossler_2019_ICCV} et al. took AUC as the core metric in the FaceForensics++ benchmark, and it has become the most widely accepted standard. Subsequently, mainstream benchmarks including Celeb-DF\cite{Li_2020_CVPR}, DFDC\cite{dolhansky2020deepfakedetectionchallengedfdc} and DFDCP\cite{dolhansky2019deepfakedetectionchallengedfdc} all adopted AUC as the primary evaluation indicator.
Although researchers have supplemented a series of targeted metrics on the basis of AUC, the current evaluation system still relies heavily on intra-dataset AUC, lacking a unified and systematic metric to accurately quantify the model's cross-domain generalization ability in real open scenarios.

\section{Methodology}

Built upon the pre-trained CLIP model as the backbone, our approach aims to improve the robustness of face forgery detection by encouraging the model to capture semantically meaningful manipulation cues rather than relying on dataset-specific artifacts. Fig.\ref{fig:structure} illustrates the overall architecture of the proposed framework. 
To achieve this goal, we introduce two key components. 
First, we propose \textbf{PaITA}, a patch-level image--text alignment module that enhances CLIP’s sensitivity to subtle manipulation artifacts through fine-grained visual–semantic alignment. 
Second, we design \textbf{FaRMoE} (Facial Region Mixture-of-Experts), which introduces region-aware expert routing to better capture organ-specific forgery patterns by leveraging specialized feature experts for different facial regions.
In the following sections, we first briefly introduce the CLIP backbone used in our framework. 
We then present the designs of PaITA and FaRMoE in detail, followed by the joint loss formulation and the overall training strategy.
\subsection{Preliminaries: Model Backbone}

Our framework is built upon the pre-trained CLIP model, which learns aligned visual–text representations from large-scale image–text data and provides strong semantic priors for visual understanding.
However, directly applying CLIP to forgery detection has two limitations. 
First, CLIP performs image–text alignment mainly at the global level, which may overlook subtle local manipulation artifacts. 
Second, the vision encoder applies a shared transformation to all patches, lacking region-aware specialization for different facial components.
To address these issues, we introduce two complementary modules. 
PaITA extends CLIP with patch-level image–text alignment to capture fine-grained manipulation cues, while FaRMoE introduces region-aware expert routing to model organ-specific forgery patterns.

\subsection{Mask-Guided Hybrid Data Augmentation}

To encourage the model to focus on manipulated facial regions during training, we introduce a mask-guided hybrid augmentation strategy that synthesizes diverse forged samples while providing explicit spatial supervision.

Given a real image $I_r$ and a fake image $I_f$, we generate a binary authenticity mask $M \in \{0,1\}^{H\times W}$ based on facial landmarks. Several facial regions (e.g., eyes, nose, mouth, or half-face) are randomly selected as manipulated regions.
Using the mask, a synthetic forged image $\tilde I$ is constructed as

\begin{equation}
\tilde I = M \odot I_f + (1-M) \odot I_r ,
\end{equation}
where $\odot$ denotes element-wise multiplication.

To further increase appearance diversity, we incorporate Self-Blended Images (SBI)\cite{Shiohara_2022_CVPR}. A transformed version of the real image $T(I_r)$ is blended with the original image using a smoothed mask $\tilde M$:

\begin{equation}
\tilde I = \tilde M \odot T(I_r) + (1-\tilde M) \odot I_r .
\end{equation}

During training, the two augmentation strategies are randomly sampled to generate diverse forged samples. The generated mask $M$ explicitly indicates manipulated regions and will later be downsampled to the patch resolution to supervise patch-level alignment.

\subsection{Facial Region Mixture-of-Experts}

\begin{figure*}[tb]
  \centering
  \includegraphics[width=1.0\linewidth]{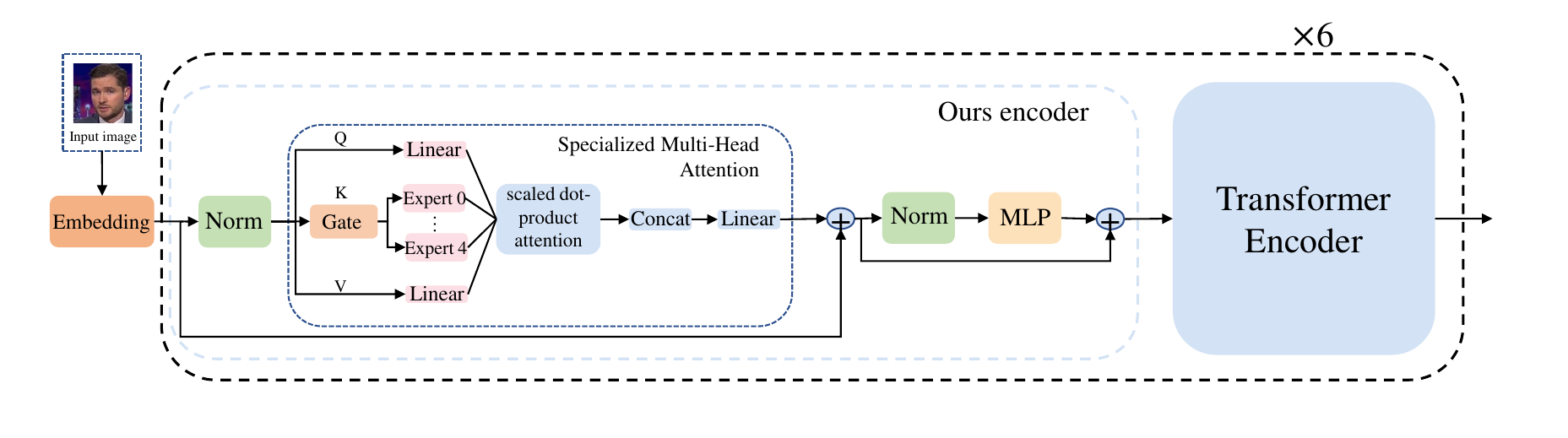}
  \caption{The structure of vision encoder with our proposed FaRMoE module.}
  \label{fig:encoder}
\end{figure*}

To capture region-specific forgery patterns, we introduce \textbf{FaRMoE}, a facial region Mixture-of-Experts module integrated into the CLIP vision encoder. Fig. \ref{fig:encoder} illustrates the structure of our targeted optimized vision encoder with \textbf{FarMoE} module. 

Given an input image, the vision encoder divides it into $N$ patches as:
$\{x_1, x_2, \dots, x_N\}, \quad x_i \in \mathbb{R}^d.$
Using facial landmarks, each patch is assigned a region label
$r_i \in \{1,\dots,K\}$,
where $K$ denotes the number of facial regions.
FaRMoE consists of a set of region experts $\{E_1,\dots,E_K\}$ that specialize in modeling different facial components. Each patch feature is routed to the corresponding expert according to its region label:

\begin{equation}
z_i = E_{r_i}(x_i).
\end{equation}

We integrate FaRMoE into the CLIP vision encoder by replacing the key projection in selected self-attention layers:

\begin{equation}
k_i = W_k x_i
\quad \rightarrow \quad
k_i = E_{r_i}(x_i).
\end{equation}

This design enables the Transformer to capture organ-specific forgery artifacts while preserving global contextual modeling.

\subsection{Patch-Level Image-Text Alignment}

To further capture subtle manipulation cues, we extend CLIP’s global image–text alignment to a patch-level visual–semantic alignment framework.
Given an input image, the vision encoder outputs a global feature $f_{cls}$ and patch features $\{f_1,f_2,\dots,f_P\}.$
On the text side, we construct two prompts describing authentic and forged content. Let $t_{real}$ and $t_{fake}$ denote their embeddings.
The visual–text similarity is computed using cosine similarity

\begin{equation}
s_{real} = \frac{f^\top t_{real}}{\|f\|\|t_{real}\|}, \quad
s_{fake} = \frac{f^\top t_{fake}}{\|f\|\|t_{fake}\|}.
\end{equation}

The forgery probability is then obtained as

\begin{equation}
p = \frac{\exp(s_{fake})}{\exp(s_{fake})+\exp(s_{real})}.
\end{equation}

Global prediction is computed from the class token $f_{cls}$, while patch-level probabilities $\{p_i\}$ are computed for each patch feature $f_i$.
The augmentation mask is downsampled to the patch resolution to obtain patch labels
$M \in \{0,1\}^{B\times P}$.

\paragraph{Intra-image local ranking loss}

To encourage higher forgery scores for manipulated patches within a forged image, we impose a ranking constraint $s_{fg}-s_{bg} \ge m$ between forged patches $S_{fg}$ and authentic patches $S_{bg}$.

The loss is defined as

\begin{equation}
\begin{split}
L_{rank\_intra} ={}&
\frac{1}{|B_{fake}|}
\sum_{b\in B_{fake}}
\frac{1}{|S_{fg}||S_{bg}|} \\
& \sum_{s_{fg}\in S_{fg}}
\sum_{s_{bg}\in S_{bg}}
\max\left(0,m-\left(s_{fg}-s_{bg}\right)\right).
\end{split}
\end{equation}

To improve generalization, we further introduce a cross-sample ranking constraint between paired real and fake samples.
For a pair $(r,f)$, let $R_{fake}$ denote forged patches in the fake image and $R_{real}$ denote the corresponding patches in the real image.

\begin{equation}
\begin{split}
L_{rank\_real\_fake} ={}&
\frac{1}{|B_{pair}|}
\sum_{(r,f)\in B_{pair}}
\frac{1}{|R_{fake}|} \\
& \sum_{i}
\max\left(0,m-\left(s_{fake}^i-s_{real}^i\right)\right).
\end{split}
\end{equation}

The overall training objective combines global classification and patch-level alignment constraints:

\begin{equation}
L_{total} =
L_{cls}
+
\lambda_1 L_{rank\_intra}
+
\lambda_2 L_{rank\_real\_fake}.
\end{equation}

Here the global classification loss is defined as

\begin{equation}
L_{cls} =
-\frac{1}{N}
\sum_{i=1}^{N}
\left[
y_i\log p_i+(1-y_i)\log(1-p_i)
\right].
\end{equation}

\section{Cross-AUC Metric For Forgery Detection}\label{sec:blind}
Area Under the Receiver Operating Characteristic Curve (AUC) is widely regarded as the dominant evaluation metric in forgery detection tasks. It quantifies a model’s discriminative ability by comparing positive and negative samples within a single dataset. However, this over-reliance on AUC overlooks a critical flaw: it only reflects model performance on pre-collected datasets, not its effectiveness in real-world detection scenarios. We argues that AUC’s dataset-bound nature leads to misleading evaluations, as it fails to account for the dynamic, complex, and diverse conditions of practical forgery detection. To evaluate cross-domain discrimination ability, we introduce \textbf{Cross-AUC}, which measures whether a detector can distinguish fake samples from one dataset against real samples from another dataset.
Let $K$ datasets be denoted as $\{D_1,\dots,D_K\}$. 
And let $R_i$ denote the set of real samples in dataset $D_i$ and $F_j$ denote the set of fake samples in dataset $D_j$. 
For each dataset pair $(i,j)$ where $i\neq j$, we compute a cross-domain AUC:

\begin{equation}
\text{AUC}_{i,j} = \text{AUC}(R_i, F_j),
\end{equation}
which measures the model's ability to distinguish real samples from domain $i$ and fake samples from domain $j$.

The final Cross-AUC score is obtained by averaging over all dataset pairs:

\begin{equation}
\text{Cross-AUC} =
\frac{1}{K(K-1)}
\sum_{i \ne j}
\text{AUC}(R_i, F_j).
\end{equation}

Compared with conventional AUC computed within a single dataset, the key contribution of the proposed Cross-AUC is to evaluate discrimination across heterogeneous domains.

\begin{table*}[tb]
\centering
  \caption{Intra-datasets comparison results with the metric of frame-level AUC. CDv1 and CDv2 denote Celeb-DF-v1 and Celeb-DF-v2 datasets, respectively.}

\begin{tabular}{ccccccc}
\hline
Model             & CDv1        & CDv2        & DFDCP       & DFDC        & UADFV       & Avg.         \\ \hline
Xception\cite{Chollet_2017_CVPR}          & 0.754 & 0.740 & 0.730 & 0.713 & 0.936 & 0.775 \\ 
Efficientnet-B4\cite{pmlr-v97-tan19a}    & 0.767 & 0.750 & 0.641 & 0.709 & 0.924 & 0.758 \\ 
F3Net\cite{10.1007/978-3-030-58610-2_6}             & 0.750 & 0.729 & 0.704 & 0.700 & 0.8910 & 0.755 \\ 
FFD\cite{Dang_2020_CVPR}               & 0.709 & 0.687 & 0.705 & 0.714 & 0.950 & 0.753 \\ 
RECCE\cite{Cao_2022_CVPR}             & 0.734 & 0.741 & 0.698 & 0.680 & 0.887 & 0.748 \\ 
UCF\cite{Yan_2023_ICCV}               & 0.811 & 0.772 & 0.693 & 0.731 & 0.920 & 0.785 \\ 
CLIP\cite{pmlr-v139-radford21a}              & 0.716 & 0.754 & 0.661 & 0.713 & 0.850 & 0.739 \\ 
Forensics Adapter\cite{Cui_2025_CVPR} & \uline{0.914} & \textbf{0.900} & \textbf{0.890} & \textbf{0.843} & \textbf{0.980} & \textbf{0.905} \\ 
Effort\cite{yan2025orthogonalsubspacedecompositiongeneralizable}            & 0.892 & 0.833 & 0.799 & 0.784 & 0.955 & 0.853 \\ \hline
\rowcolor{tabgray}
Ours              & \textbf{0.916} & \uline{0.885} & \uline{0.862} & \uline{0.796} & \uline{0.972} & \uline{0.886} \\ \hline
\end{tabular}
\label{table 1}

\end{table*}

\begin{table*}[tb]
\centering
  \caption{Intra-datasets comparison results with the metric of video-level AUC. CDv1 and CDv2 denote Celeb-DF-v1 and Celeb-DF-v2 datasets, respectively.}
\begin{tabular}{ccccccc}
\hline
Model             & CDv1        & CDv2        & DFDCP       & DFDC        & UADFV       & Avg.         \\ \hline
Xception\cite{Chollet_2017_CVPR}          & 0.810 & 0.816 & 0.761 & 0.740 & 0.963 & 0.818 \\ 
Efficientnet-B4\cite{pmlr-v97-tan19a}    & 0.815 & 0.808 & 0.675 & 0.724 & 0.960 & 0.796 \\ 
F3Net\cite{10.1007/978-3-030-58610-2_6}            & 0.811 & 0.789 & 0.735 & 0.718 & 0.930 & 0.797 \\ 
FFD\cite{Dang_2020_CVPR}              & 0.761 & 0.742 & 0.741 & 0.739 & 0.973 & 0.791 \\ 
RECCE\cite{Cao_2022_CVPR}             & 0.815 & 0.823 & 0.715 & 0.696 & 0.943 & 0.798 \\ 
UCF\cite{Yan_2023_ICCV}               & 0.861 & 0.837 & 0.706 & 0.751 & 0.955 & 0.822 \\ \
CLIP\cite{pmlr-v139-radford21a}              & 0.770 & 0.814 & 0.679 & 0.736 & 0.895 & 0.779 \\ 
Forensics Adapter\cite{Cui_2025_CVPR} & \textbf{0.969} & \textbf{0.957} & \textbf{0.929} & \uline{0.872} & \textbf{0.994} & \textbf{0.944} \\ 
Effort\cite{yan2025orthogonalsubspacedecompositiongeneralizable}            & 0.935 & 0.885 & 0.816 & 0.807 & 0.965 & 0.882 \\ \hline
\rowcolor{tabgray}
Ours              & \uline{0.966} & \uline{0.950} & \uline{0.900} & \textbf{0.898} & \uline{0.984} & \uline{0.940} \\ \hline
\end{tabular}
\label{table 2}

\end{table*}
\section{Experiments}
We conduct comprehensive and systematic experiments to validate the effectiveness of our proposed \textbf{SFAM} deepfake detection framework and the novel \textbf{Cross-AUC} evaluation metric. We first detail the complete experimental setup, including datasets, evaluation metrics, and implementation details. Then, we present the cross-domain performance comparison with state-of-the-art (SOTA) methods, followed by ablation studies to quantify the independent contribution of each core component in our framework.

\subsection{Experimental Settings}

\subsubsection{Datasets}
To evaluate the performance of our proposed forgery detection method, we select a representative training dataset and five public test datasets, covering various forgery techniques and data distributions to ensure the generalization ability of the model. The details of the datasets are described as follows. FaceForensics++\cite{Rossler_2019_ICCV} is adopted as the training dataset, which is one of the most widely used benchmark datasets for face forgery detection. And five public test datasets, including Celeb-DF-v1, Celeb-DF-v2\cite{Li_2020_CVPR}, DFDCP\cite{dolhansky2019deepfakedetectionchallengedfdc}, DFDC\cite{dolhansky2020deepfakedetectionchallengedfdc} and UADFV\cite{8630787} are used to evaluate the generalization performance of the testing model, covering different forgery methods, including NeuralTextures\cite{10.1145/3306346.3323035}, Face2Face\cite{Thies_2016_CVPR}, FaceSwap, FaceShifter and Deepfakes.

\subsubsection{Evaluate Metrics}
We adopt four core metrics to comprehensively evaluate the performance of all models, which are fully aligned with the research objectives and can systematically reflect the model's detection accuracy, cross-domain generalization ability, and practical deployment stability:
(1) \textit{AUC}: Traditional intra-dataset AUC, calculated by pairing real and fake samples within a single test dataset. This metric is used for horizontal comparison with existing methods and to verify the basic detection capability of the model in closed-set scenarios.
(2) \textit{Standard Deviation (Std)}: The standard deviation of 20 cross-domain Cross-AUC values, which measures the fluctuation of the model's performance across different cross-domain scenarios. A smaller Std indicates more stable and reliable performance of the model in diverse real-world environments.
(3) \textit{Cross-AUC}: The value obtained from all dataset pairs, which is the core metric to evaluate the overall cross-domain generalization ability of the model in practical scenarios.
(4) \textit{Cross-AUC Minimum}: The minimum value among all testing Cross-AUC, which reflects the model's detection performance in the most challenging cross-domain scenario, and is a key indicator to evaluate the lower bound of the model's practical deployment performance.

\subsubsection{Implementation Details}
We adopt a two-stage training strategy consistent with the framework design. First, we pre-train the \textbf{PaIMA} module based on the original CLIP model for 10 epochs. Then, we replace 6 key (k) networks in the multi-head self-attention layers of the pre-trained CLIP ViT encoder with our proposed \textbf{FaRMoE} module. With the pre-trained model keeping frozen, we fine-tune the new structure for 1 additional epoch until convergence. All experiments are conducted on two Nvidia 3090 GPUs.

\subsection{Comparison Results}

\subsubsection{Intra-dataset Comparison Experiment}

Table \ref{table 1} and Table \ref{table 2} present the frame-level and video-level AUC results of all compared models on five widely-used public benchmark datasets. 
From the overall results, the Forensics Adapter model achieves the highest average frame-level AUC of 0.905. Our proposed SFAM framework maintains a highly competitive average frame-level AUC of 0.886, which is on par with top-tier state-of-the-art (SOTA) methods and exhibits strong fundamental detection capability in intra-dataset scenarios.

Notably, our model delivers outstanding performance in video-level detection, with an average video-level AUC of 0.940. This result is only 0.4 percentage points lower than Forensics Adapter (0.944), and significantly outperforms other mainstream baselines. Specifically, we reach the highest frame-level AUC of 0.916 on the Celeb-DF-v1 dataset, surpassing Forensics Adapter (0.914) and all other compared methods. On the most challenging DFDC dataset, which features diverse real-world forgery patterns, complex background noise, and heterogeneous data distributions, our model achieves the highest video-level AUC of 0.898, which is 2.6 percentage points higher than Forensics Adapter (0.872). This demonstrates that our framework maintains stable and reliable detection performance across consecutive video frames, which is a critical capability for practical real-world video deepfake detection tasks.

\subsubsection{Cross-dataset Comparison Experiment}
We conduct cross-domain detection experiments on the five datasets for the SFAM framework, as well as all baseline models. The experimental results are summarized in Table \ref{tab:cross_dataset_full}.
\begin{table*}[tb]
  \centering
  \setlength{\tabcolsep}{1pt}
  \caption{Cross-datasets comparison results with several metrics. CDv1 and CDv2 denote Celeb-DF-v1 and Celeb-DF-v2 datasets, respectively.}

  \scalebox{1.0}{

    \begin{tabular}{ccccccccccc>{\columncolor{tabgray}}c} 
      \hline
       & & Xception & Efficientnet-B4 & F3Net & FFD & RECCE & UCF & CLIP & Forensics Adapter & Effort & Ours \\
      \hline
      \multirow{4}{*}{\makecell[c]{CDv1}}
      & CDv2   & 0.787 & 0.794 & 0.789 & 0.792 & 0.766 & 0.845 & 0.766 & \textbf{0.945} & 0.902 & 0.929 \\ \hhline{~~----------} 
      & DFDC   & 0.803 & 0.747 & 0.721 & 0.684 & 0.700 & 0.805 & 0.627 & \textbf{0.986} & 0.923 & 0.938 \\ \hhline{~~----------}
      & DFDCP  & 0.818 & 0.812 & 0.768 & 0.723 & 0.746 & 0.800 & 0.714 & \textbf{0.980} & 0.893 & 0.932 \\ \hhline{~~----------}
      & UADFV  & 0.804 & 0.788 & 0.761 & 0.775 & 0.722 & 0.819 & 0.739 & \textbf{0.960} & 0.908 & 0.946 \\
      \hline
      \multirow{4}{*}{\makecell[c]{CDv2}}
      & CDv1   & 0.697 & 0.713 & 0.677 & 0.571 & 0.704 & 0.720 & 0.697 & 0.857 & 0.815 & \textbf{0.865} \\ \hhline{~~----------}
      & DFDC   & 0.764 & 0.692 & 0.643 & 0.547 & 0.670 & 0.719 & 0.601 & \textbf{0.969} & 0.871 & 0.900 \\ \hhline{~~----------}
      & DFDCP  & 0.779 & 0.771 & 0.705 & 0.588 & 0.719 & 0.709 & 0.696 & \textbf{0.959} & 0.821 & 0.892 \\ \hhline{~~----------}
      & UADFV  & 0.763 & 0.745 & 0.696 & 0.666 & 0.691 & 0.738 & 0.722 & \textbf{0.923} & 0.845 & 0.912 \\
      \hline
      \multirow{4}{*}{\makecell[c]{DFDC}}
      & CDv1   & 0.660 & 0.659 & 0.737 & 0.735 & 0.730 & 0.691 & 0.760 & 0.688 & 0.731 & \textbf{0.814} \\ \hhline{~~----------}
      & CDv2   & 0.703 & 0.700 & 0.781 & 0.828 & 0.763 & 0.743 & 0.810 & 0.750 & 0.749 & \textbf{0.837} \\ \hhline{~~----------}
      & DFDCP  & 0.746 & 0.726 & 0.754 & 0.752 & 0.744 & 0.681 & 0.749 & \textbf{0.872} & 0.738 & 0.854 \\ \hhline{~~----------}
      & UADFV  & 0.731 & 0.699 & 0.748 & 0.810 & 0.719 & 0.712 & 0.775 & 0.786 & 0.770 & \textbf{0.875} \\
      \hline
      \multirow{4}{*}{\makecell[c]{DFDCP}}
      & CDv1   & 0.623 & 0.645 & 0.673 & 0.699 & 0.656 & 0.742 & 0.712 & 0.665 & 0.778 & \textbf{0.747} \\ \hhline{~~----------}
      & CDv2   & 0.664 & 0.680 & 0.715 & 0.791 & 0.691 & 0.788 & 0.757 & 0.724 & 0.794 & \textbf{0.775} \\ \hhline{~~----------}
      & DFDC   & 0.695 & 0.629 & 0.643 & 0.674 & 0.631 & 0.740 & 0.636 & \textbf{0.859} & 0.834 & 0.805 \\ \hhline{~~----------}
      & UADFV  & 0.698 & 0.683 & 0.691 & 0.775 & 0.650 & 0.757 & 0.735 & 0.754 & 0.810 & \textbf{0.824} \\
      \hline
      \multirow{4}{*}{\makecell[c]{UADFV}}
      & CDv1   & 0.922 & 0.923 & 0.891 & 0.926 & 0.903 & 0.923 & 0.842 & 0.960 & 0.951 & \textbf{0.961} \\ \hhline{~~----------}
      & CDv2   & 0.935 & 0.934 & 0.911 & 0.959 & 0.917 & 0.940 & 0.875 & \textbf{0.975} & 0.952 & 0.965 \\ \hhline{~~----------}
      & DFDC   & 0.936 & 0.907 & 0.871 & 0.908 & 0.864 & 0.912 & 0.764 & \textbf{0.992} & 0.959 & 0.968 \\ \hhline{~~----------}
      & DFDCP  & 0.944 & 0.937 & 0.891 & 0.934 & 0.894 & 0.912 & 0.827 & \textbf{0.989} & 0.949 & 0.966 \\
      \hline
      \multicolumn{2}{c}{AUC Avg.}    & 0.775 & 0.758 & 0.753 & 0.748 & 0.734 & 0.785 & 0.739 & \textbf{0.905} & 0.853 & 0.886 \\ \hhline{~~----------}
      \multicolumn{2}{c}{Cross-AUC Avg.} & 0.774 & 0.759 & 0.753 & 0.757 & 0.744 & 0.785 & 0.740 & 0.881 & 0.850 & \textbf{0.885} \\ \hhline{~~----------}
      \multicolumn{2}{c}{Cross-AUC Min}         & 0.623 & 0.629 & 0.643 & 0.547 & 0.631 & 0.681 & 0.601 & 0.665 & 0.731 & \textbf{0.747} \\ \hhline{~~----------}
      \multicolumn{2}{c}{Std}         & 0.094 & 0.097 & 0.080 & 0.115 & 0.083 & 0.081 & 0.069 & 0.102 & 0.074 & \textbf{0.066} \\
      \hline
    \end{tabular}
  }
\label{tab:cross_dataset_full}
\end{table*}

\textit{Verification of the Limitations of Traditional AUC}.
The experimental results provide solid evidence for our core argument about the limitations of traditional AUC in Section 1. The Forensics Adapter model achieves the highest Original AUC Average of 0.905, indicating its excellent closed-set intra-dataset detection performance. However, this high intra-dataset AUC does not translate to reliable cross-domain generalization: its Cross-AUC Minimum drops to only 0.665, which is 12.3 percentage points lower than our SFAM model (0.747), and its performance fluctuation Std reaches 0.102, 54.5\% higher than ours (0.066). This contrast directly validates that traditional AUC is inherently dataset-bound, and models optimized solely for intra-dataset AUC tend to overfit to dataset-specific forgery patterns rather than learning universal forgery features. In contrast, our SFAM model maintains a negligible gap of only 0.1 percentage points between Original AUC Average (0.886) and Cross-AUC Average (0.885), demonstrating that it learns generalizable facial forgery details rather than dataset-specific biases. 
This result proves that the proposed Cross-AUC metric reflects the real-world detection performance of models more accurately, and serves as a more practical and comprehensive evaluation standard for deepfake detection research.

\textit{Superiority in Cross-Domain Generalization}.
Our proposed model achieves the highest Cross-AUC Average of 0.885 among all evaluated models, outperforming all mainstream and state-of-the-art baselines. This performance demonstrates that our SFAM framework achieves remarkable cross-domain generalization ability for deepfake detection. It effectively addresses the core challenge of existing methods, which suffer from severe performance degradation when facing unseen forgery patterns and heterogeneous data in practical scenarios.

\textit{Deployment Stability Across Real-World Scenarios}.
Our SFAM framework achieves the highest Cross-AUC Minimum of 0.747 among all evaluated models. This result proves that our model maintains reliable and effective detection performance while effectively avoiding the catastrophic performance degradation that widely exists in baseline methods. 

Meanwhile, our model reaches an extremely low Cross-AUC Std of 0.066, which is the lowest among all methods. This demonstrates that our model has consistent and stable detection performance across diverse data distributions and forgery patterns, without drastic performance fluctuations. The outstanding performance in both two core stability indicators fully validates that our SFAM framework has far more practical deployment value than existing methods for real-world deepfake detection.

\begin{table*}[t]
  \centering
  \caption{Ablation study of different components in the proposed SFAM framework. MGHDA denotes the Mask-Guided Hybrid Data Augment module, PaITA denotes the patch-level Image-Text Alignment, and FaRMoE denotes the Facial Region Mixture-of-Experts module.}
  \label{tab:ablation}
  \begin{tabular}{ccccccc}
    \hline
     & MGHDA & PaITA & FaRMoE & AUC Avg. & Cross-AUC Avg. & Cross-AUC Min \\
    \hline
    & $\times$ & $\times$ & $\times$ & 0.739 & 0.740 & 0.601 \\
    \hline
    & $\checkmark$ & $\times$ & $\times$ & 0.858 & 0.861 & 0.725 \\
    \hline
    & $\checkmark$ & $\checkmark$ & $\times$ & 0.880 & 0.881 & \textbf{0.749} \\
    \hline
    & $\checkmark$ & $\checkmark$ & $\checkmark$ & \textbf{0.886} & \textbf{0.885} & 0.747 \\
    \hline
  \end{tabular}
\end{table*}

\subsection{Ablation Study}
We first conduct ablation experiments on the three core components of our SFAM framework to explore the effect of each component. The experimental results are shown in Table \ref{tab:ablation}. Key observations from the ablation results are summarized as follows:
First, the CLIP baseline achieves the worst performance across all metrics, with a Cross-AUC Average of only 0.740 and a Cross-AUC Minimum of 0.601. This demonstrates that the CLIP model lacks optimization and cannot be directly applied to deepfake detection without targeted structural and training strategy modifications.
Second, mask-guided hybrid data augmentation serves as the foundation for performance improvement. By adding this module, the Cross-AUC Average increases significantly from 0.740 to 0.861, and the Cross-AUC Minimum rises from 0.601 to 0.725. This improvement validates that the mask-guided hybrid data augmentation enriches the diversity of fake samples, and explicitly guides the model to focus on forged regions, thus fundamentally enhancing the model's cross-domain generalization ability.
Third, patch-level alignment is the core module for further cross-domain performance enhancement. On the basis of mask-based augmentation, adding patch-level alignment further lifts the Cross-AUC Average to 0.881 and Cross-AUC Minimum to 0.749. This result proves that the fine-grained cross-modal semantic matching between local image patches and regional text prompts enables the model to capture subtle local forgery cues that are overlooked by global-only alignment.

\begin{table*}[t]
\centering
  \caption{Experiment results of different hyperparameters in the loss function.}

\begin{tabular}{ccccc}
\hline
        & $\lambda_1$     & $\lambda_2$ & AUC Avg. & Cross-AUC Avg. \\ \cline{1-5}
        & 0.2       & 0.3               & 0.831            & 0.833 \\ \cline{1-5}
        & 0.25       & 0.25            & 0.868            & 0.865 \\ \cline{1-5}
        & 0.3       & 0.2              & \textbf{0.880}            & \textbf{0.881} \\ \cline{1-5}
        & 0.35       & 0.15             & 0.865           & 0.866 \\ \cline{1-5}
\end{tabular}
\label{table para}
\end{table*}

Finally, the full SFAM model with all three modules achieves the highest Original AUC Average of 0.886, while reaching a highest Cross-AUC Average of 0.885 and Cross-AUC Minimum of 0.747, which is nearly identical to Variant 2. This indicates that the \textbf{FaRMoE} module further enhances the model's organ-specific fine-grained feature extraction and overall detection ability, without sacrificing cross-domain generalization performance or the lower bound of detection stability.
In summary, the three core modules have a significant effect, and their addition continuously improves the model's detection accuracy, cross-domain generalization ability, and practical deployment stability. All ablation results fully validate the rationality and necessity of each module's design in our SFAM framework.

\subsection{Parameter Analysis}
We also conduct experiments on the hyperparameters $\lambda_1$ and $\lambda_2$ of the loss function in section 3.4 to explore the best configuration. The experiment is conducted without the \textbf{FaRMoE} module and the results are shown in Table \ref{table para}.
The results show that the combination of ($\lambda_1$=0.3 and $\lambda_2$=0.2) achieves the highest Original AUC Average (0.880) and Cross-AUC Average (0.881) among all tested configurations. 
The intra-image ranking loss $L_{rank\_intra}$ (weighted by $\lambda_1$) guides the model to distinguish forged and authentic regions within a single image, which is the foundation of detection ability. We assign a slightly higher weight of 0.3 to ensure the model can capture subtle local forgery cues. The cross-sample ranking loss $L_{rank\_real\_fake}$ (weighted by $\lambda_2$) helps the model learn universal forgery features and improve cross-domain generalization, so we set a weight of 0.2 to provide effective supervision without affecting the basic feature learning.
The weight setting of ($\lambda_1$=0.3 and $\lambda_2$=0.2) achieves the best balance between the two ranking loss terms. All other weight configurations lead to obvious performance degradation. For configuration of ($\lambda_1$=0.2 and $\lambda_2$=0.3), the reversed weight allocation makes the model ignore fine-grained local forgery details, resulting in a sharp drop in overall performance. The equal weight setting fails to capture the different priorities of the two loss terms, and cannot reach the optimal performance. For the last configuration of ($\lambda_1$=0.35 and $\lambda_2$=0.15), excessive weight on intra-image loss makes the model overfit to local details of the training set, which damages the cross-domain generalization ability.

Besides, the optimal configuration has a negligible gap between Original AUC and Cross-AUC, which means the model achieves the best balance between closed-set detection performance and real-world cross-domain generalization. This perfectly matches the core design goal of our SFAM framework.

\begin{figure*}[tb]
  \centering
  \includegraphics[width=0.8\linewidth]{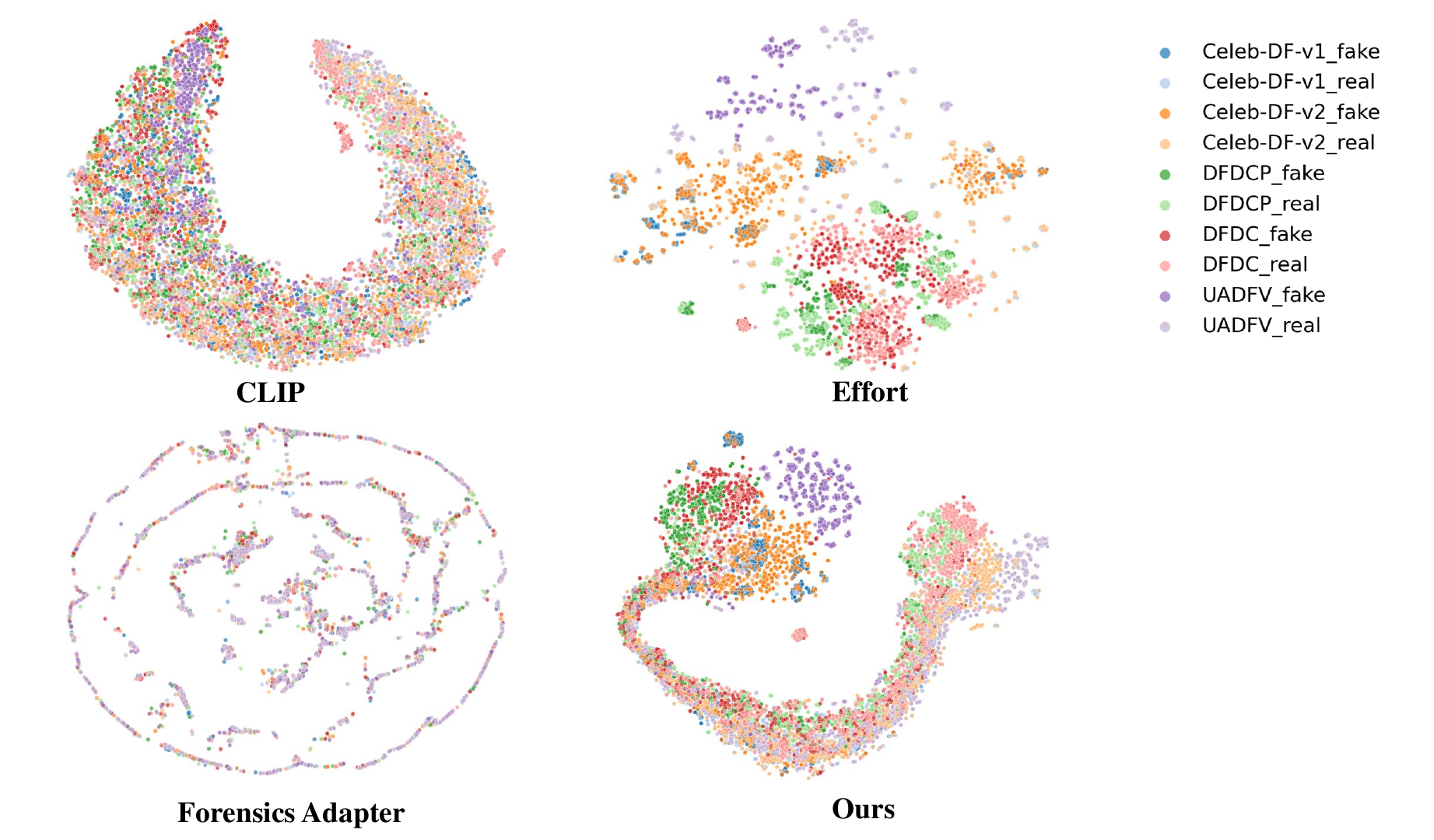}
  \caption{t-SNE visualization results of four methods.}
  \label{fig:visualization}
\end{figure*}

\subsection{Visualization Analysis}
We perform t-distributed Stochastic Neighbor Embedding (t-SNE)\cite{hinton_visualizing_2008} visualization on the feature representations learned by four methods: the CLIP baseline, Effort, Forensics Adapter, and our proposed SFAM framework, with the results presented in Figure \ref{fig:visualization}.
As observed from the figure, the features of real and forged images extracted by the CLIP baseline are severely intertwined with no clear discriminative boundary between the two categories. This indicates that the baseline fails to effectively distinguish between authentic and fake samples.
In contrast, Effort and Forensics Adapter form relatively well-defined class-wise distributions in the feature space. However, the features extracted by these methods exhibit obvious inter-dataset distribution gaps: samples from different data sources form isolated clusters in the feature space. This phenomenon reveals that the models rely to dataset-specific biases, which may impair their generalization ability in real-world open scenarios.

Meanwhile, our proposed method exhibits superior cross-dataset generalization capability. Samples from different data sources are distributed more compactly in the feature space, while the clear separation between authentic and fake categories is well preserved. This demonstrates that our framework effectively mitigates the adverse effects of dataset bias and learns more generalizable forgery-related feature representations, thereby enhancing the model's robustness in complex real-world deployment scenarios.

\section{Conclusion}
In this paper, we have proposed a novel Cross-AUC evaluation metric for generalizable forgery detection task in the real.
The proposed Cross-AUC provides a more practical evaluation of real-world generalization performance.
Besides, we further proposed the novel framework Semantic Fine-grained Alignment and Mixture-of-Experts (SFAM), consisting of a patch-level image-text alignment module that enhances CLIP’s sensitivity to manipulation artifacts, and the facial region mixture-of-experts module, which routes features from different facial regions to specialized experts for region-aware forgery analysis.
Extensive experiments validate that our method achieves state-of-the-art cross-domain detection performance on mainstream benchmarks.
In the future, we will evaluate the proposed method with different forgery models on more complex real scenarios to adapt to the needs of the real world.



%
%
\bibliographystyle{splncs04}
\bibliography{references}
\end{document}